%% file: 2020-icip-parallax-pipeline.tex
\documentclass{article}

\usepackage{spconf,amsmath,graphicx}
\usepackage{amssymb}
\usepackage{colortbl}
\usepackage{url}
\usepackage{color}
\usepackage[table]{xcolor}
\usepackage{booktabs}
\usepackage{multirow}
\usepackage{subcaption}
\usepackage[sort,nocompress]{cite}
\usepackage{stfloats}
\usepackage{textcomp}
\usepackage[pdfa]{hyperref}
\makeatletter
\newcommand*\titleheader[1]{\gdef\@titleheader{#1}}
\AtBeginDocument{%
  \let\st@red@title\@title
  \def\@title{%
    \vskip-5.5em
    \bgroup\normalfont\footnotesize\centering\@titleheader\par\egroup
    \vskip1.5em\st@red@title}
}
\makeatother

\newcommand{\review}[1]{\textcolor{black}{#1}}

\title{Parallax Motion Effect Generation \\ through Instance Segmentation and Depth Estimation}

\titleheader{
\copyright 2020 IEEE. Personal use of this material is permitted. Permission from IEEE must be obtained for all other uses, in any current or future media, including reprinting/republishing this material for advertising or promotional purposes, creating new collective works, for resale or redistribution to servers or lists, or reuse of any copyrighted component of this work in other works. Pre-print of article that will appear in 2020 IEEE International Conference on Image Processing (ICIP), Abu Dhabi, United Arab Emirates, 2020, pp. 1621-1625, doi: 10.1109/ICIP40778.2020.9191168. \\
The published article is available on \url{https://ieeexplore.ieee.org/document/9191168}
}

\name{%
  \begin{tabular}{@{}c@{}}
    Allan Pinto$^1$ \,
    Manuel A. C\'{o}rdova$^1$ \,
    Luis G. L. Decker$^1$ \,
    Jose L. Flores-Campana$^1$ \,
    Marcos R. Souza$^1$ \\
    Andreza A. dos Santos$^1$ \,
    Jhonatas S. Concei\c{c}\~{a}o$^1$ \,
    Henrique F. Gagliardi$^2$ \\
    Diogo C. Luvizon$^2$ \,
    Ricardo da S. Torres$^3$ \, 
    Helio Pedrini$^1$
  \end{tabular}%
  \thanks{We thank Samsung R\&D Institute Brazil for the financial support. This work was funded by Samsung Eletr\^{o}nica da Amaz\^{o}nia Ltda., through the project ``Parallax Effect'', within the scope of the Informatics Law No. 8248/91. Authors are grateful to Coordena\c{c}\~{a}o de Aperfei\c{c}oamento de Pessoal de N\'{i}vel Superior -- CAPES (Finance Code 001), National Council for Scientific and Technological Development -- CNPq (grant \#309330/2018-1) and S\~ao Paulo Research Foundation -- FAPESP (grants \#2016/50250-1, \#2017/12646-3 and \#2019/16253-1).}
}

\address{
    $^1$Institute of Computing, University of Campinas (UNICAMP), Campinas, SP, Brazil, 13083-852 \\
    $^2$AI R\&D Lab, Samsung R\&D Institute Brazil, Campinas, SP, 13097-160, Brazil\\
    $^3$NTNU -- Norwegian University of Science and Technology, {\AA}lesund, Norway.
}

\begin{document}

\ninept
\tolerance=999
\sloppy

\maketitle

\input{00-abstract.tex}
\input{01-introduction.tex}
\input{02-proposed-method.tex}
\input{03-experiments-results.tex}

\input{04-conclusions.tex}

\bibliographystyle{IEEEbib}
\bibliography{bib/IEEEabrv,bib/strings,bib/refs,bib/references}

\end{document}

%% file: 00-abstract.tex
\begin{abstract}
Stereo vision is a growing topic in computer vision due to the innumerable opportunities and applications this technology offers for the development of modern solutions, such as virtual and augmented reality applications. \review{To enhance the user's experience in three-dimensional virtual environments, the motion parallax estimation is a promising technique to achieve this objective.} In this paper, we propose an algorithm for generating parallax motion effects from a single image, taking advantage of state-of-the-art instance segmentation and depth estimation approaches. This work also presents a comparison against such algorithms to investigate the trade-off between efficiency and quality of the parallax motion effects, taking into consideration a multi-task learning network capable of estimating instance segmentation and depth estimation at once. Experimental results and visual quality assessment indicate that the PyD-Net network (depth estimation) combined with Mask R-CNN or FBNet networks (instance segmentation) \review{can} produce parallax motion effects with good visual quality.
\end{abstract}


%% file: 01-introduction.tex
\section{Introduction}
\label{sec:introduction}

\review{Stereo vision~\cite{kim2019fast,li2018stereo}} is a growing topic in computer vision (CV) due to the innumerable opportunities this technology offers for developing modern applications, such as virtual and augmented reality \review{systems~\cite{Okura2016ISMAR,Pathak2017SII}}, entertainment~\cite{Thatte2016ICME}, autonomous robot navigation~\cite{Prucksakorn2018RAS}, and medicine~\cite{Liao2010TBME}. While common tasks and problems (e.g., image classification) in computer vision are concerned with the development of algorithms for identifying, understanding and analyzing 2D images, the stereo vision and 3D reconstruction tasks aim the design of models and algorithms able to infer 3D properties from objects presented in a scene and then reconstruct the spatial relationship between them.

\begin{figure}[!t]
\centering
\includegraphics[width=0.95\columnwidth]{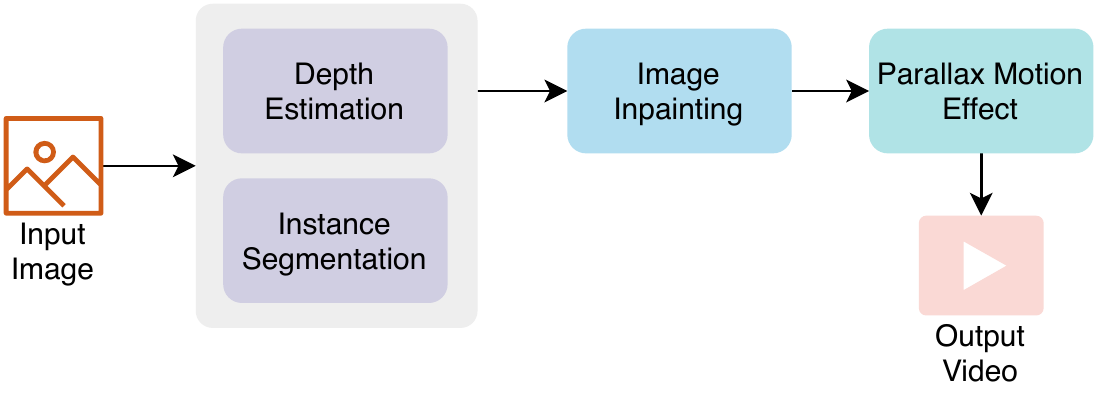}
\caption{Overview of the proposed methodology for generating parallax motion effect from images.}
\label{fig:project-pipeline}
\end{figure}

Although several advances have been reported in the literature for 3D reconstruction and stereo vision problems, the understanding of 3D information in a scene from images is still an open challenge due mainly to inherent ill-posing nature of estimating depth information from pixels based on their intensity values~\cite{Choi2015TIP}. To overcome these limitations, the use of machine learning techniques and data from 3D sensors has been proposed to minimize the errors in the inference related to ambiguity issues during 3D reconstruction~\cite{shahnewaz2020color,Choi2015TIP,Poggi2018IROS}. In this context, CV research community has spent efforts to provide good quality data from 3D imaging sensor to enable accurate CV-based machine learning models for some specific tasks, such as autonomous driving. Examples of good quality data for this task include Cityscapes and KITTI datasets~\cite{Cordts2016CVPR,Menze2018ISPRS}.

Among the techniques available for 3D reconstruction and visualization, depth estimation in binocular vision systems, disparity estimation from stereo images and motion parallax estimation from a sequence of images are certainly the most promising techniques for achieving this objective~\cite{cai2020light,yoon2020depth,hamzah2020depth}. In particular, the spatial perception stimulus generated by motion parallax has propelled several theoretical studies in the areas of visual perception and psychology that seek physiological explanations in humans, towards establishing the neurological bases of our visual ability~\cite{Rogers1979Perception,Ono1986JEPHPP,Stroyan2012JMB}.

Motion parallax~\cite{mansour2019relative,kim2016neural,zhang2019parallax,layton2020computational,serrano2019motion} provides an important monocular depth cue raised from the relative velocity between the objects and the observer. In motion parallax, objects near the observer move faster than objects that are farther away. Since this motion is considered a rich source of 3D information~\cite{Rogers2016Perception,Schindler2016NI}, several computer vision studies have recently proposed the use of motion parallax to enrich depth perception in tasks involving 3D scene reconstructions~\cite{Jones2008VRC,Thatte2016ICME,Kellnhofer2016SIGGRAPH}.

Based on evidences that motion parallax can potentially enrich human depth visual perception, this research aims to devise algorithms and methods to automatically generate motion parallax effect from a single image, in order to provide a 3D immersion experience to the user with devices equipped with a general-purpose RGB camera. Since there is no good quality dataset available for this task, this work aims to answer the following research questions: (i) Could CV-based machine learning models, originally proposed for depth estimation problem, be adapted to generate motion parallax effects, with a good visual quality? (ii) Are the state-of-the-art methods for instance segmentation able to generalize enough to enable their use in scenarios whose image acquisition is different from those considered in the training time? To answer these questions, we proposed a method for parallax motion effect that takes advantage of recent developments for instance segmentation and depth estimation problems, as illustrated in Fig.~\ref{fig:project-pipeline}.

The remaining of this text is organized as follows. Section~\ref{sec:proposed-method} introduces the proposed method for motion parallax effect generation. Section~\ref{sec:experiments-results} presents and discusses the achieved results. Finally, Section~\ref{sec:conclusions} provides our conclusions and future research venues.

%% file: 02-proposed-method.tex
\section{Proposed Method}
\label{sec:proposed-method}

This section presents the proposed method to generate a video considering the use of parallax motion concepts to move objects in an image. The proposed method was designed to produce parallax motions, considering three types of movements: zoom in, the left and right. Regardless of the movement type considered, we propose the use of a simple speed model to determine the relative position of the foreground and background components at a given instant $t$. The following sections discuss the main steps of our method.

\vspace*{0.1cm}
\noindent
\textbf{Merging the Results of Instance Segmentation and Depth Estimations Networks.}
This step aimed to join the results from instance segmentation and depth estimation methods to capture the scene semantic context associated with spatial relations among the objects in the scene. First, we used an instance segmentation algorithm to find the boundary of objects in an image $I$. Next, we applied a depth estimation method to find the position of these objects on the $z$-axis. Finally, we sorted them to get the nearest object to the camera.

\begin{figure}[!t]
    \centering
    \begin{subfigure}[c]{0.30\columnwidth}
        \centering
        \includegraphics[width=0.99\columnwidth]{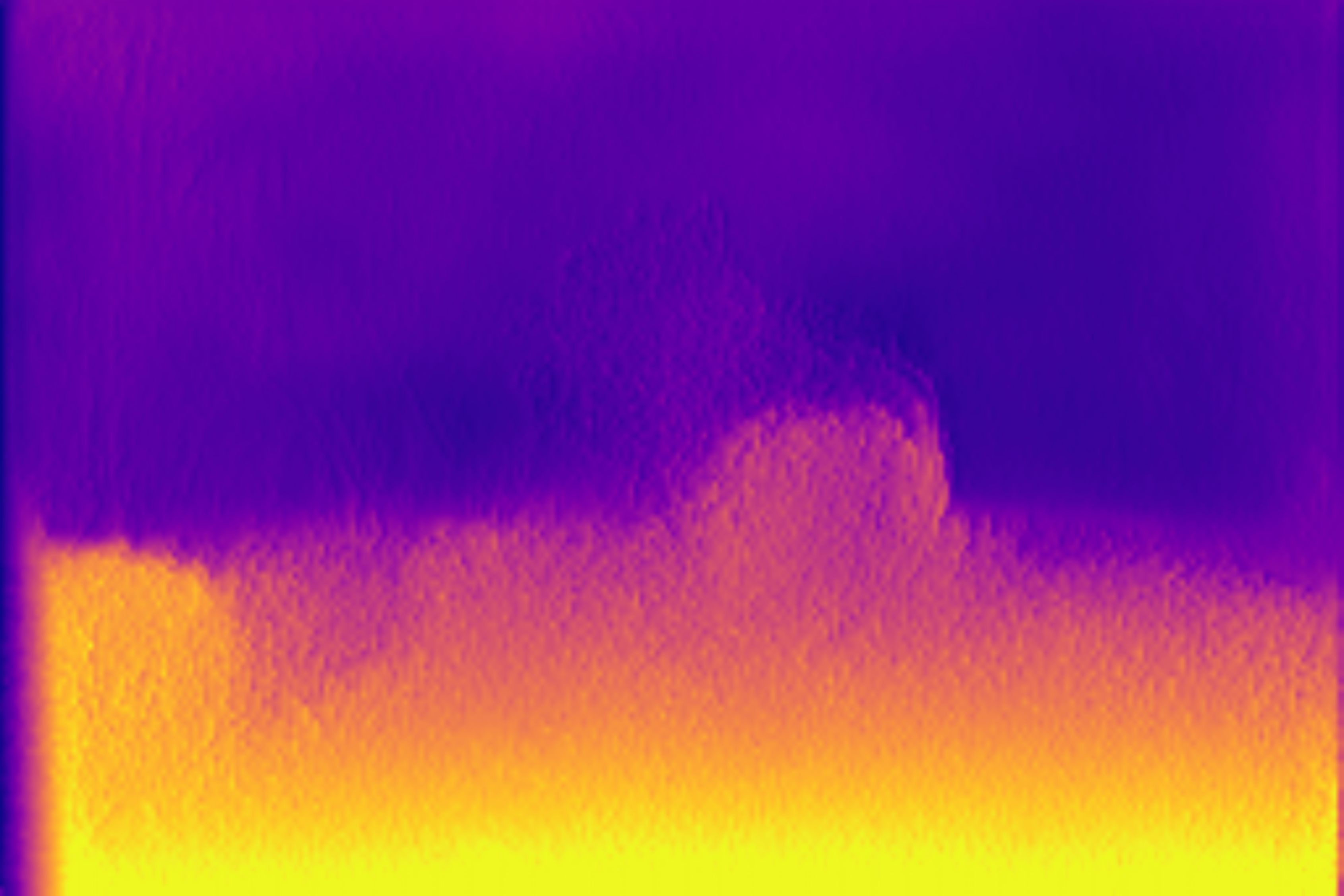}
        \caption{Depth map.}
    \end{subfigure}
    \begin{subfigure}[c]{0.30\columnwidth}
        \centering
        \includegraphics[width=0.99\columnwidth]{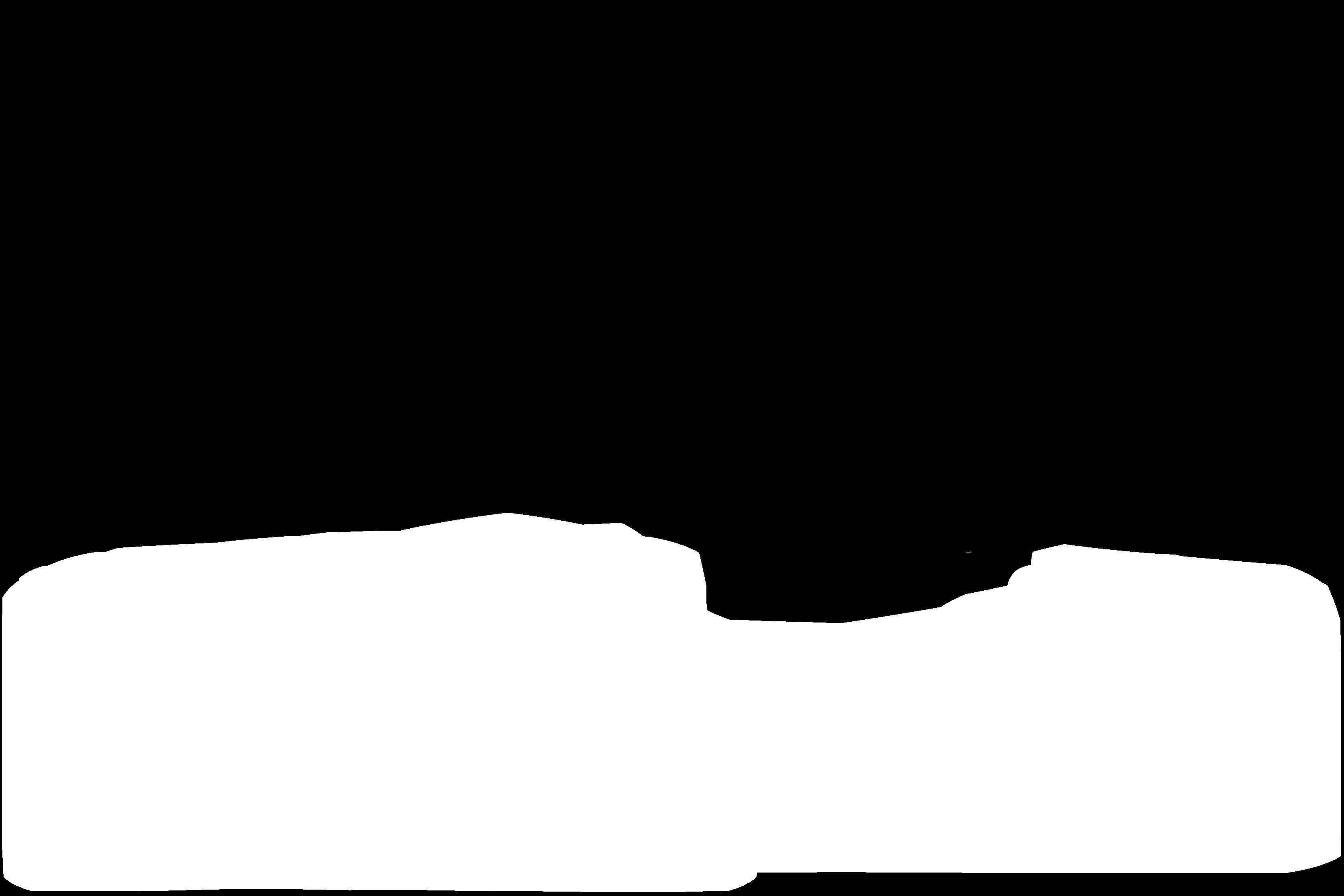}
        \caption{Nearest object.}
    \end{subfigure}
    \begin{subfigure}[c]{0.30\columnwidth}
        \centering
        \includegraphics[width=0.99\columnwidth]{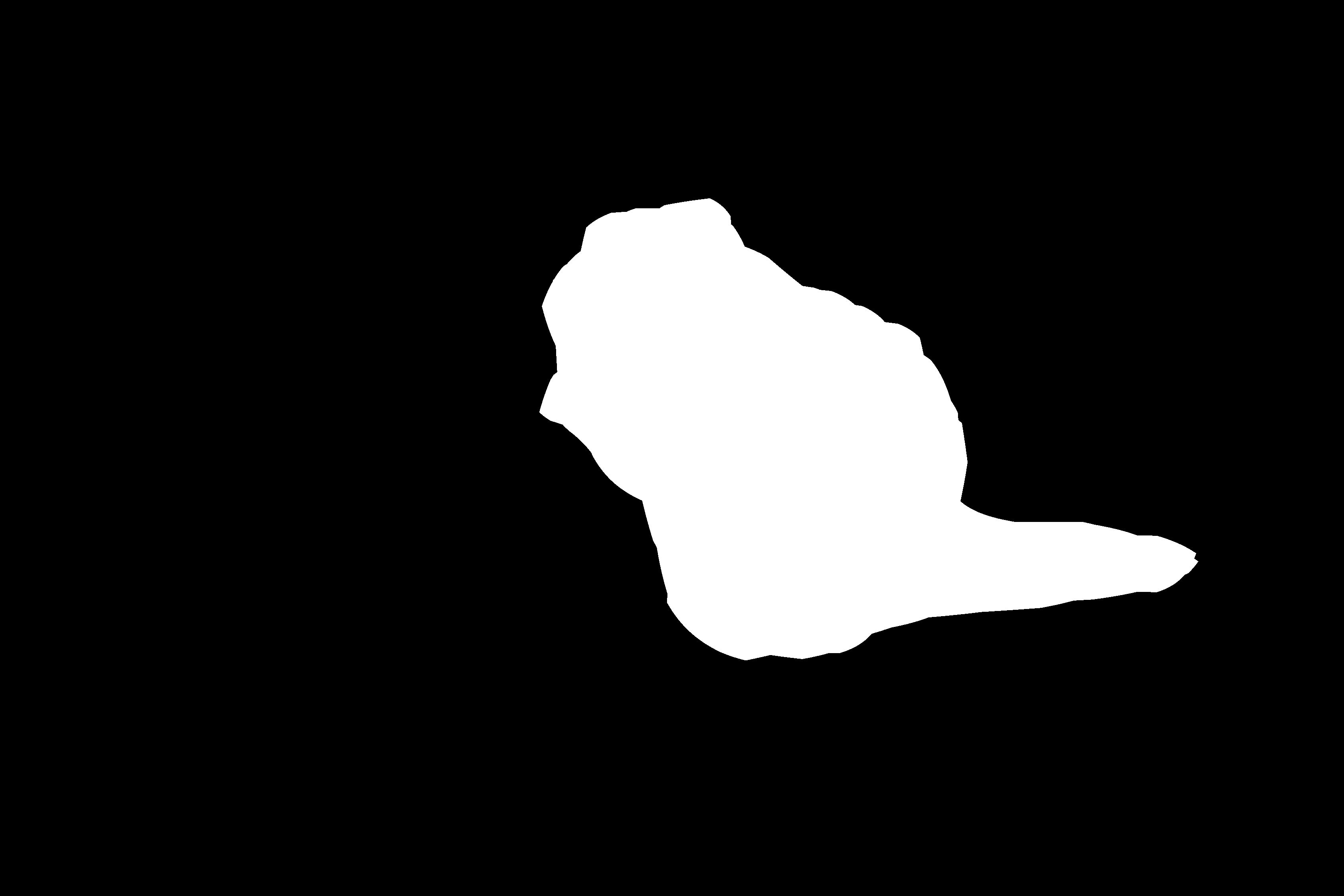}
        \caption{\review{Farthest object.}}
    \end{subfigure}
    \caption{Example of detected objects (squirrel and stone) sorted according to the average depth value around their center of mass.}
    \label{fig:example_step_1}
\end{figure}

To sort the objects according to their $z$-axis positions, firstly, we computed a binary mask for all segmented objects. Next, we used these masks to compute the center of mass of the objects. Finally, we averaged the depth values considering a $5\times5$ kernel size around the center of mass, which were used to sort the objects' masks (see Fig.~\ref{fig:example_step_1}). The mask with the highest disparity value (nearest to the camera) was used to: (i) isolate the nearest object, which was clipped and pasted into a new image with a transparent background, hereafter named as foreground component; and (ii) to remove the nearest object from the original image to produce a new image without the foreground object, hereafter named as background component.

\vspace*{0.1cm}
\noindent
\textbf{Refinement of Background and Foreground Components and Image Inpainting.}
After finding the \textit{nearest mask}, which is used to produce the background and foreground components, the next step aims to perform a post-processing upon this mask to remove erroneous pixels in both background and foreground components, caused by segmentation errors. In summary, this step is essential: (i) to prevent that the inpainting method fills out the holes in the background image using the objects' pixels left in the image, after the object removal; and (ii) to enhance the boundary of the objects that comprise the foreground component by removing pixels belonging to background.

\begin{figure}[!t]
    \centering
    \begin{subfigure}[c]{0.70\columnwidth}
        \centering
        \includegraphics[width=0.44\columnwidth]{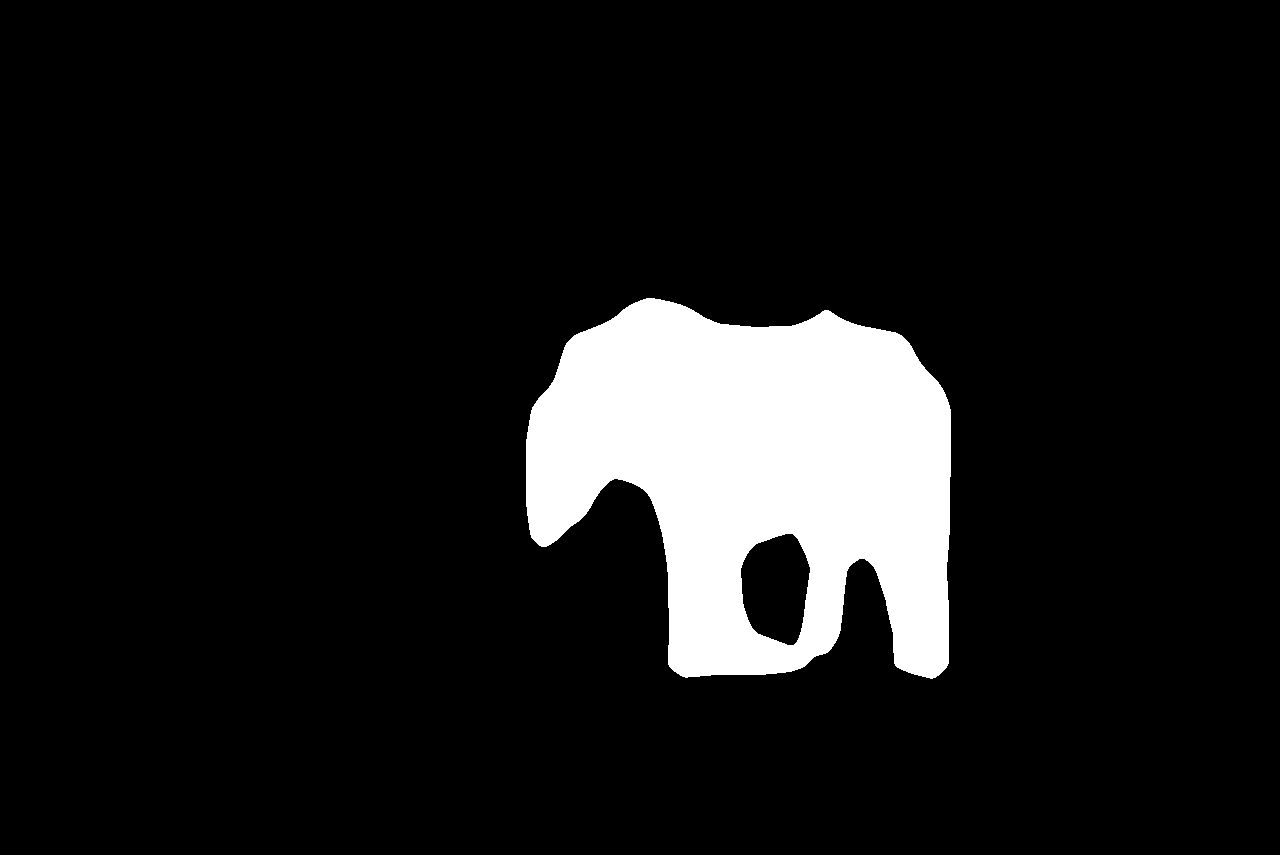}
        \includegraphics[width=0.44\columnwidth]{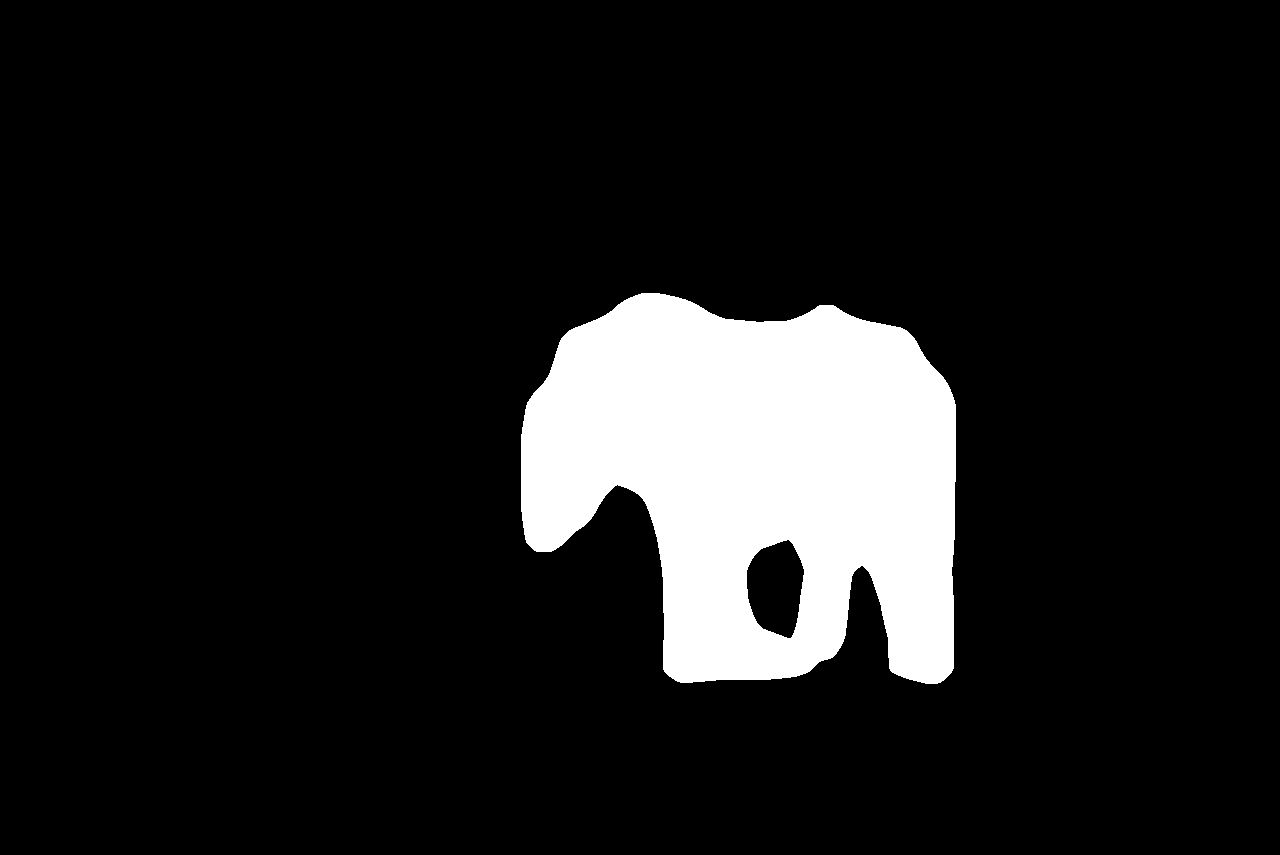}
    \end{subfigure}
    \begin{subfigure}[c]{0.70\columnwidth}
        \centering
        \includegraphics[width=0.44\columnwidth]{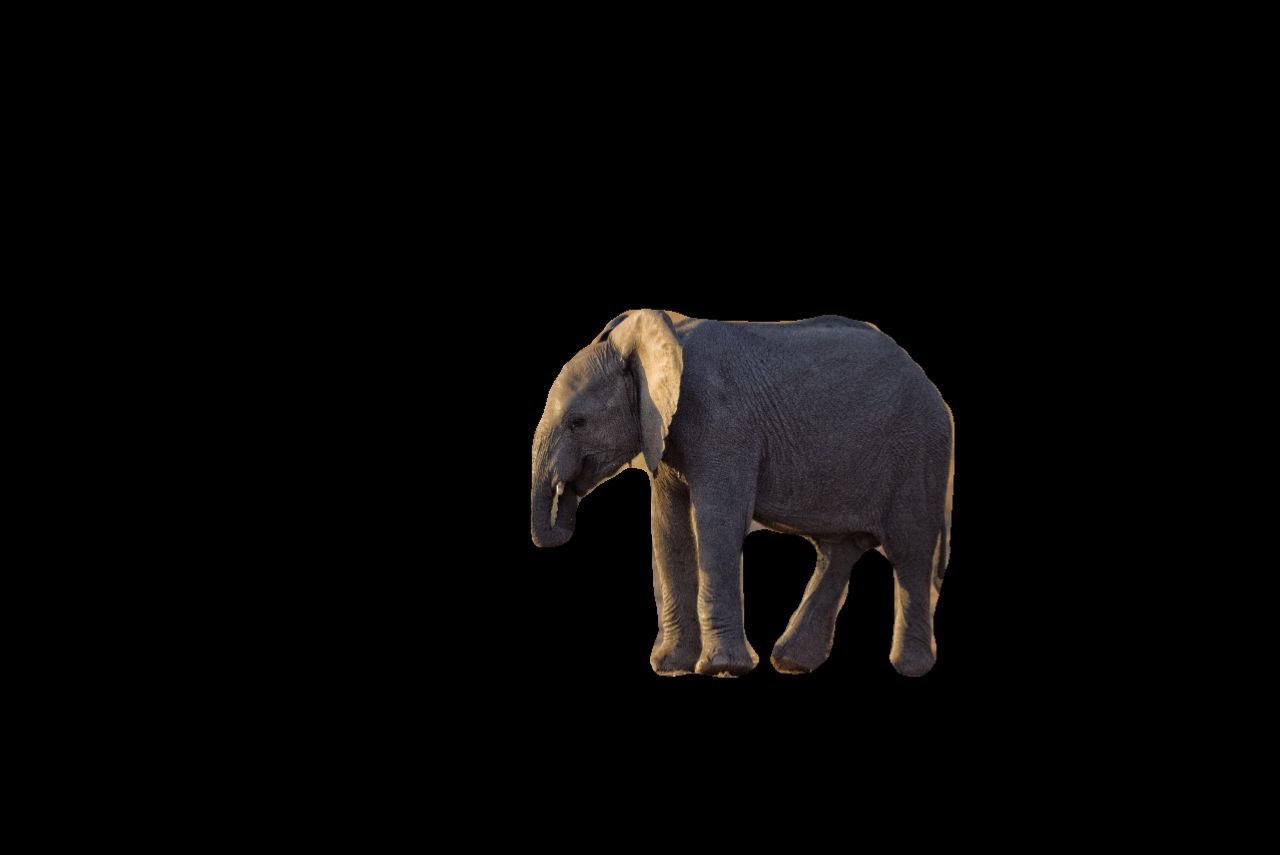}
        \includegraphics[width=0.44\columnwidth]{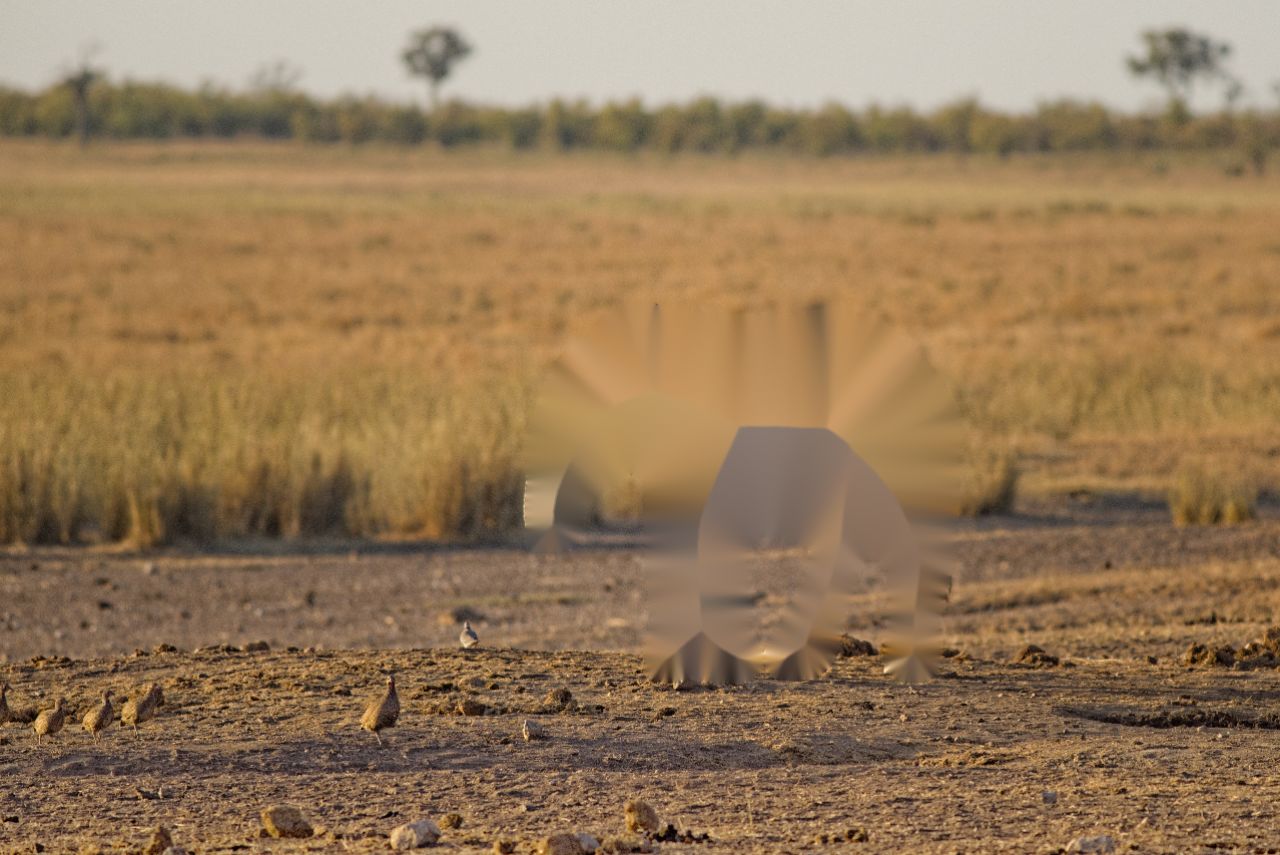}
    \end{subfigure}
    \caption{Example of refined masks (first row) and background and foreground components (second row). The top-left image illustrates a refined mask used to produce a foreground component (bottom left), while the bottom-left image shows a refined mask used to produce the background component (bottom right).}
    \label{fig:example_step_inpainting}
\end{figure}

To refine the foreground component, first, we applied a Gaussian blur, considering a kernel size of 7$\times$7 upon the \textit{nearest mask}. Next, we threshold the smoothed mask to come up with a new one, which was used to produce the refined foreground component. On the other hand, to refine the background component, we performed a dilation of the nearest mask considering a kernel size of $11\times11$ towards enlarging the region of interest coded into the binary mask, and thus come up with a coarse object's delimitation to ensure removal of all pixels that belong to the object. Finally, we used the Telea's~\cite{Telea2004GGT} inpainting algorithm to fill out the hole left in the image after the object removal (see Fig.~\ref{fig:example_step_inpainting}).

\vspace*{0.1cm}
\noindent
\textbf{Speed Model for Background and Foreground Components.}
The parallax motion is simulated through a simple and efficient method to compute the movement of the background and foreground components. According to the concepts of motion parallax, the object near to the camera moves faster than objects far from the camera. In this initial solution, we simulated this effect by considering the use of finite arithmetic sequences, with $n$ elements, for both components but with different constant terms, \review{as shown in Eq.~\ref{eq:speed_model}}:
\begin{align}
    \textit{fore}_{n} = \textit{fore}_{1} + (n-1) \times c_{\textit{fore}} \nonumber \\
    \textit{back}_{n} = \textit{back}_{1} + (n-1) \times c_{\textit{back}}
    \label{eq:speed_model}
\end{align}
\noindent where $\textit{fore}_1$ and $\textit{back}_1$ are the foreground and background components used to produce the $1$-st frame of a video containing parallax motion effects, $\textit{fore}_n$ and $\textit{back}_n$ are the foreground and background components, respectively, used to produce the $n$-th frame, and the coefficients $c_{\textit{fore}}$ and $c_{\textit{back}}$ are constant terms that defines the speed movement. \review{In this context, each value of these sequences is used as a sum factor to compute the 2D geometric transformation of the background and foreground component, regardless the movement type. In such circumstances, small constant terms produce movements slower than movements produced with larger constant terms. As a result of this process, we ended up with $n$ background and $n$ foreground images, which were blended to generate a video clip containing the parallax motion effect.}

\begin{figure}[!t]
    \centering
    \begin{subfigure}[c]{0.70\columnwidth}
        \centering
        \includegraphics[width=0.44\columnwidth]{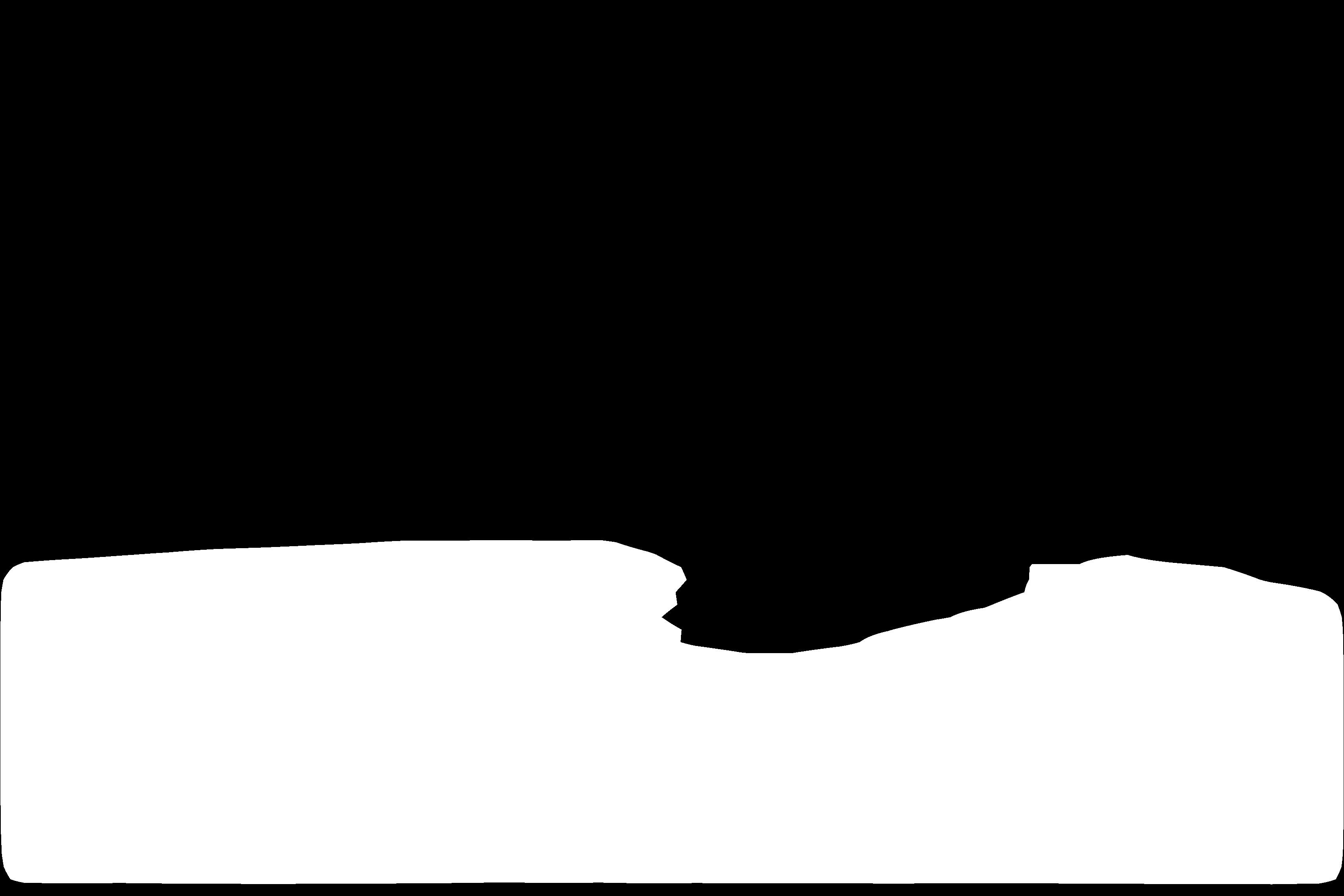}
        \includegraphics[width=0.44\columnwidth]{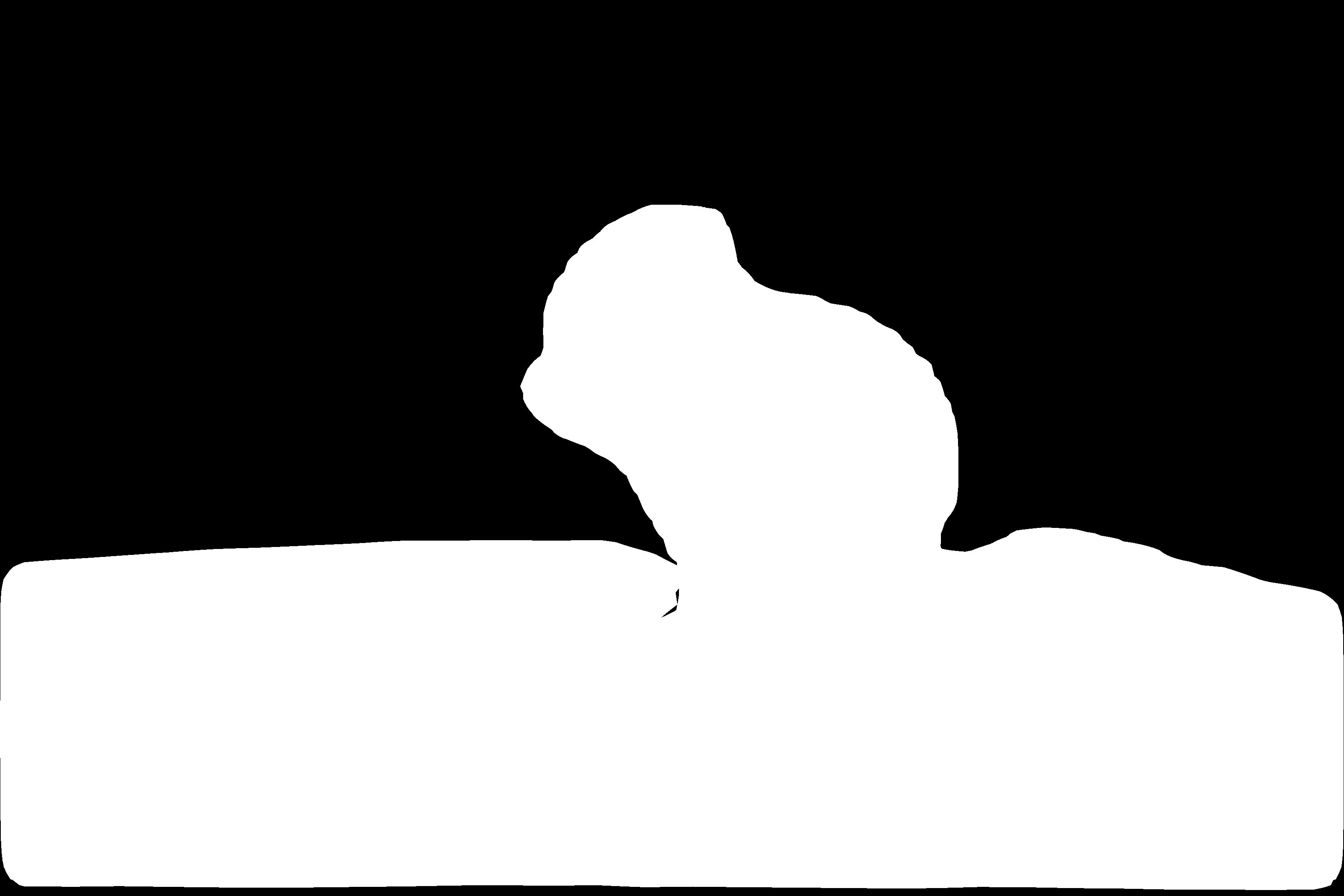}
    \end{subfigure}
    \caption{Example of a background image without any post-processing (left image) and its refined version (right image).}
    \label{fig:example_refined_background}
\end{figure}

\vspace*{0.1cm}
\noindent
\textbf{Enhancing the Quality of Parallax Motion Generation.}
We adopted three strategies to enhance the visual quality of parallax motion effects, as follows:

\begin{itemize}

\item \textbf{Small object filtering.}
This step aims to filter out small objects that are irrelevant to the parallax motion effect generation. The criterion adopted to define the minimum size of the objects in the image corresponds to relative area of objects, compared to the area of the largest object in the image. All objects with a relative area smaller than $5\%$ are added to the background layer.

\item \textbf{Joining near objects.}
To mitigate the effect of possible depth estimation errors, we devised an algorithm to join near objects considering a relative tolerance between their distance. After computing the average of depth values for each segmented objects, we sorted the objects according to their distances and then we joined pair of objects with a relative distance up to $20\%$. This strategy is useful to generate parallax motion effect for images without a clear object of interest.

\item \textbf{Two-layered scene.}
We also proposed a procedure to join objects from different classes, but that should be in the foreground component. Fig.~\ref{fig:example_refined_background} \review{illustrates} an example in which the squirrel and stone should be considered as a foreground component. However, due to bad depth estimation, both ``objects'' are far apart from each other. \review{To overcome this problem}, we slice the scene into two layers, according to the median value of depth values. In this context, an object is classified as being of the background layer if the average of their depth value is smaller than median value of the whole depth map. Otherwise, the object is classified as being of the foreground layer.

\end{itemize}

%% file: 03-experiments-results.tex
\section{Experiments and Results}
\label{sec:experiments-results}

This section presents the datasets and evaluation protocols used to validate the proposed method. We report the quality of obtained results considering metrics adopted in each category of algorithms used in this work, i.e., instance segmentation and depth estimation.

\subsection{Datasets and Metrics}

In this section, we briefly describe the datasets and evaluation protocols adopted in this work to validate our method.

\vspace*{0.1cm}
\noindent
\textbf{COCO 2017 Dataset.}
This dataset was proposed to be used in three tasks in the COCO 2017 Place Challenge: scene parsing, scene instance segmentation, and semantic boundary detection~\cite{Zhou2017CVPR}. In this work, we used the data available for the scene instance segmentation, which aims to segment an image into object instances.

\vspace*{0.1cm}
\noindent
\textbf{KITTI 2015 Dataset.}
The KITTI dataset~\cite{Menze2018ISPRS} was built considering an autonomous driving platform equipped with several acquisition sensors for collecting a wide gamma of information including stereo images (grayscale and color), optical flow estimations, visual odometry, 3D points estimations, geographic localization, among others.

\begin{figure}[!t]
\centering
    \begin{subfigure}[c]{0.35\textwidth}
        \centering
        \includegraphics[height=0.075\textheight]{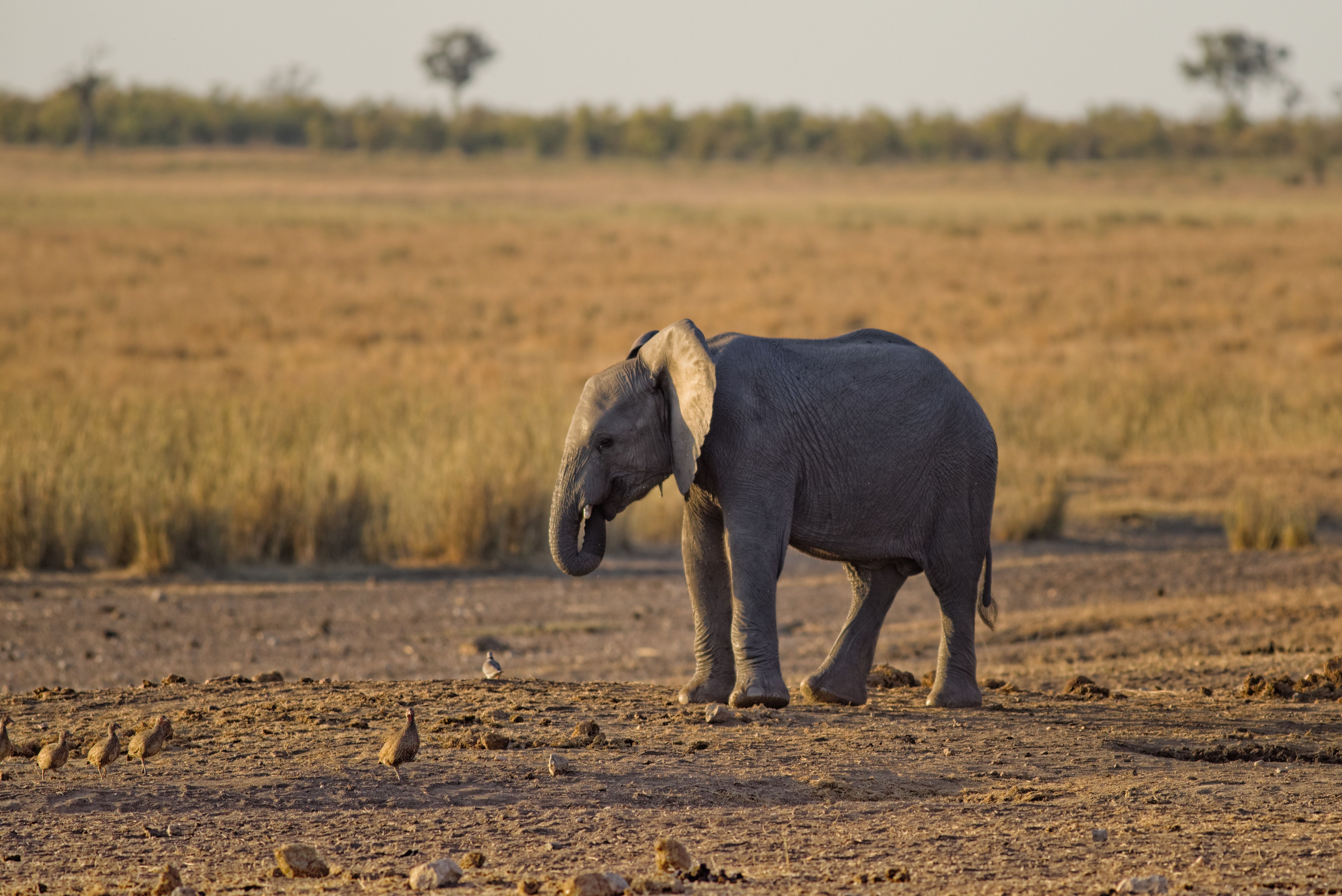}
        \includegraphics[height=0.075\textheight]{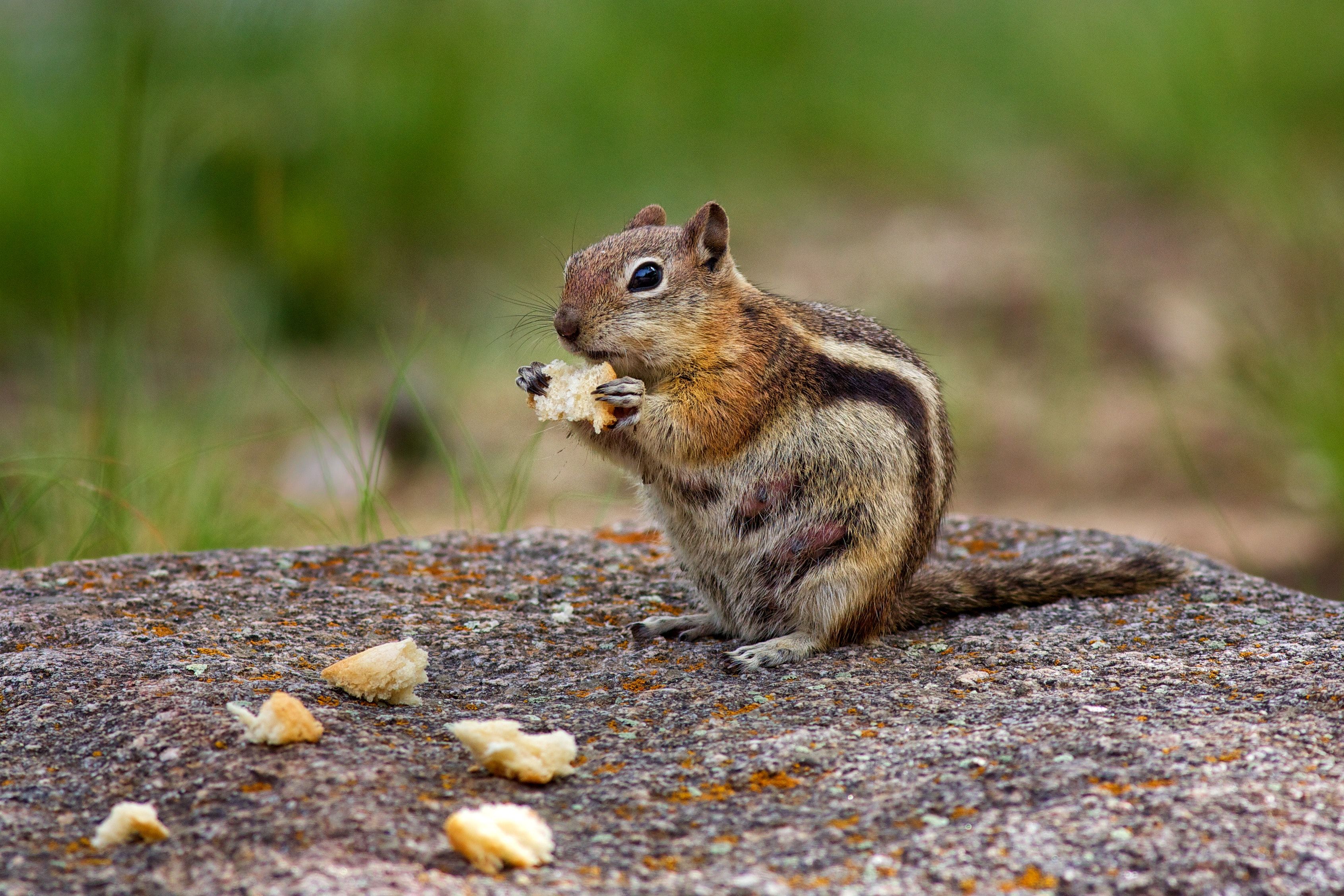}
    \end{subfigure}
    \begin{subfigure}[c]{0.35\textwidth}
        \centering
        \includegraphics[height=0.075\textheight]{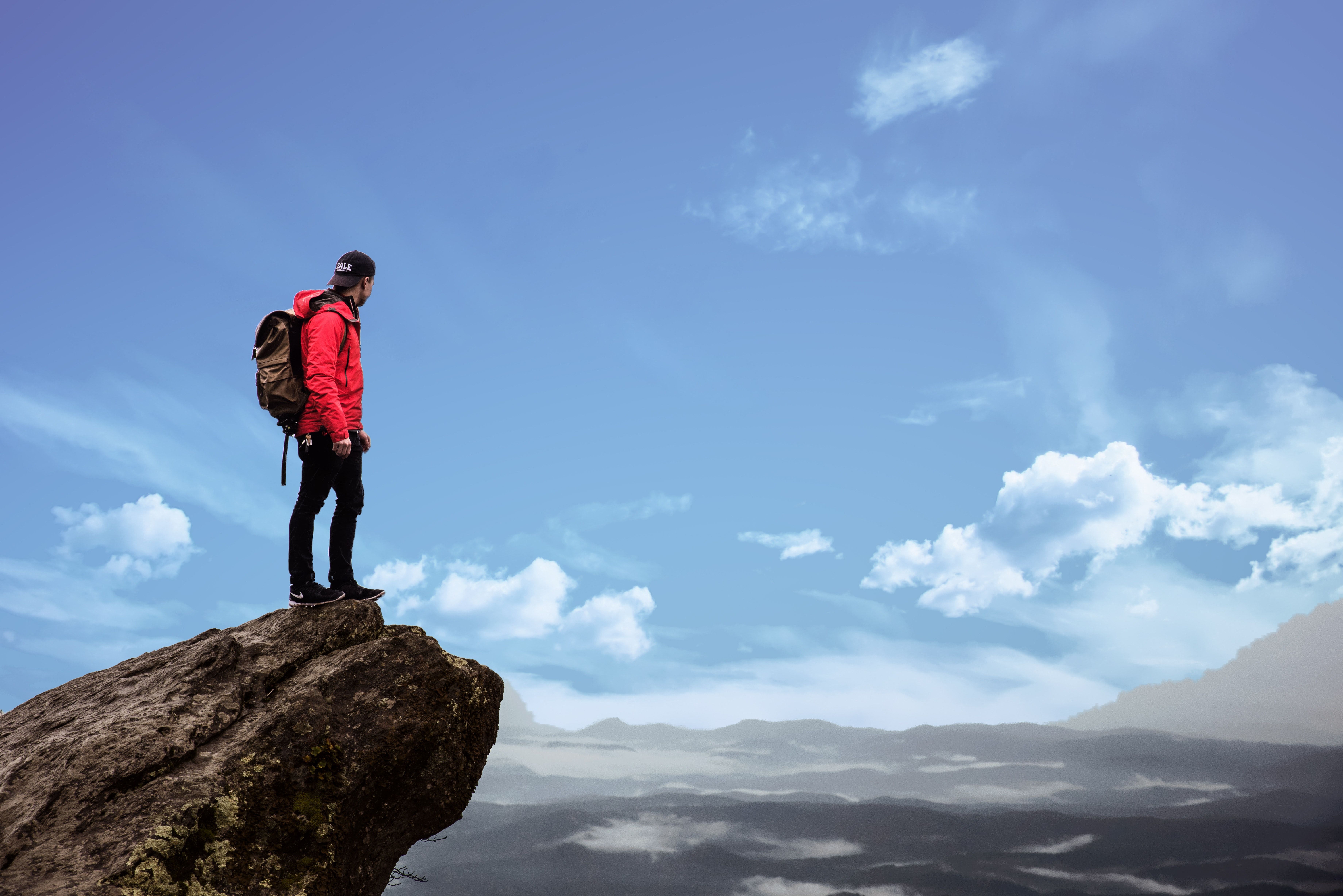}
        \includegraphics[height=0.075\textheight]{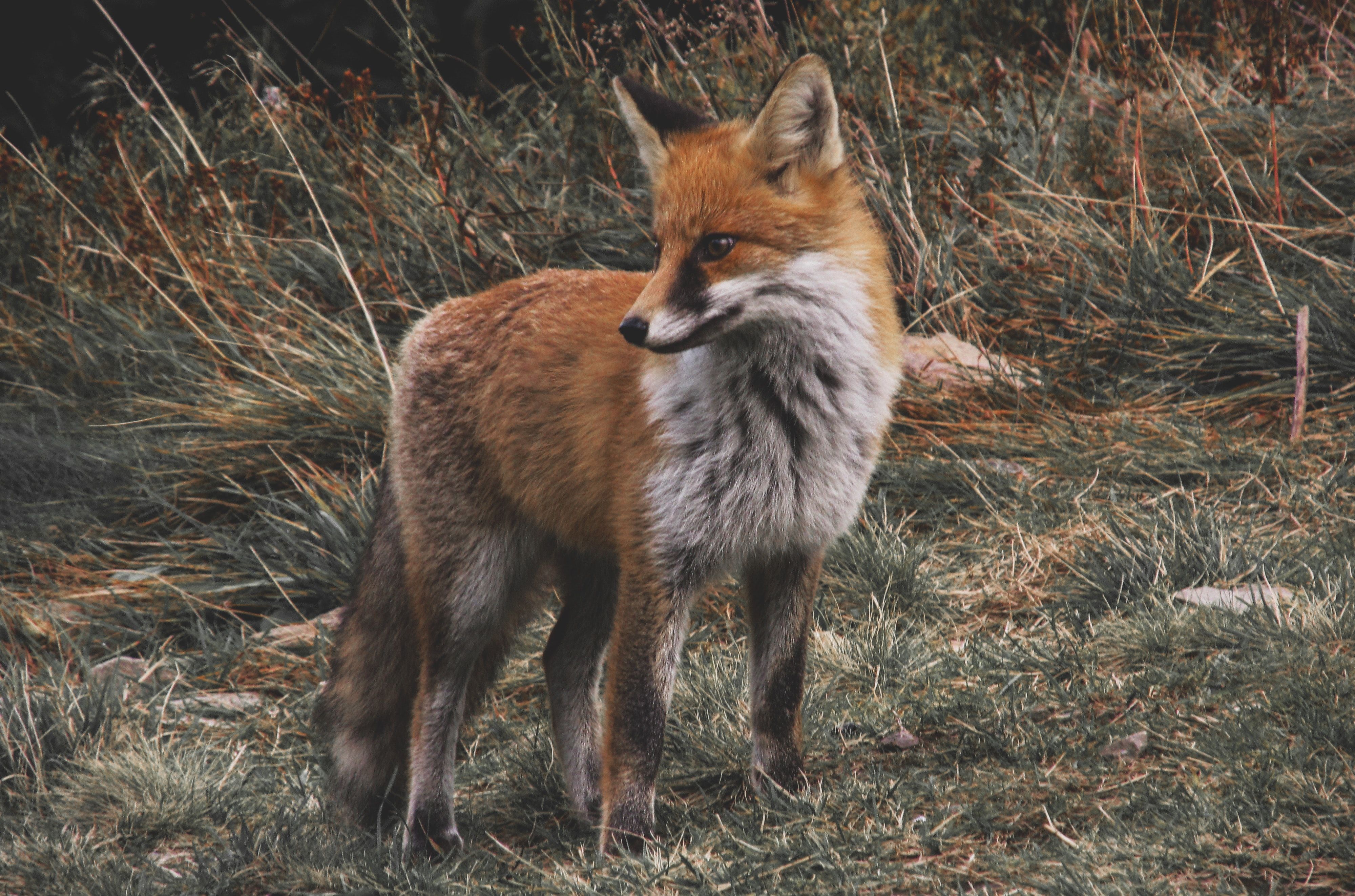}
    \end{subfigure}

\caption{Examples of images from the Parallax60 dataset.}
\label{fig:google-images}
\end{figure}

\vspace*{0.1cm}
\noindent
\textbf{Parallax60 Dataset.}
This dataset contains sixty images collected over Internet, which comprises high-quality and ultra-high-definition (UHD) images (from $3,840 \times 2,160$ to $8,192 \times 5,461$) with different backgrounds \review{(see Fig.~\ref{fig:google-images}). Most of the images are natural scenes with various types of vegetation, which makes this dataset the hardest one to generate parallax motion effect.}

\begin{figure}[!t]
\centering
    \begin{subfigure}[c]{0.90\columnwidth}
        \centering
        \includegraphics[width=0.30\textwidth]{figs/datasets-examples/2a99d71c.jpg}
        \includegraphics[width=0.30\textwidth]{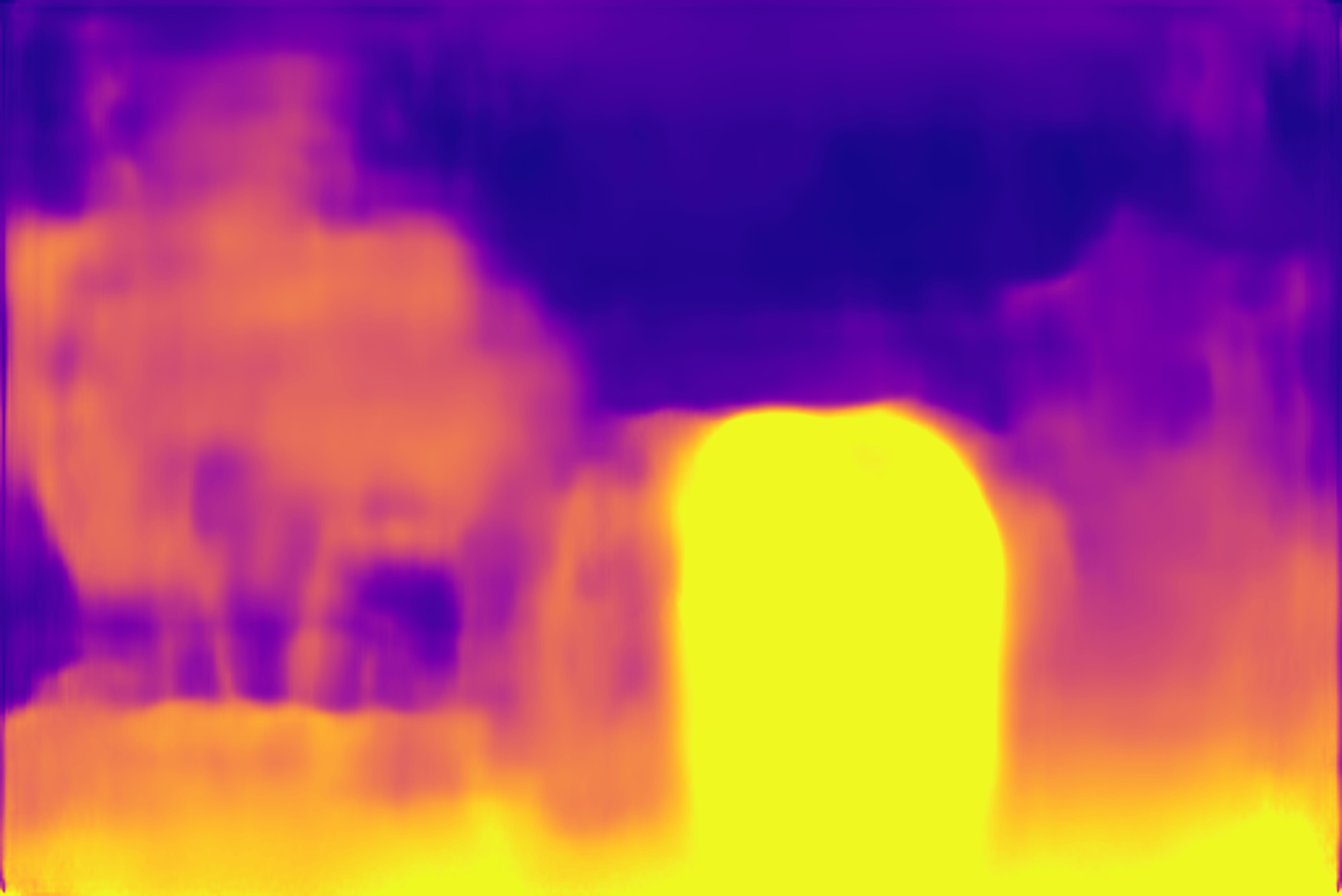}
        \includegraphics[width=0.30\textwidth]{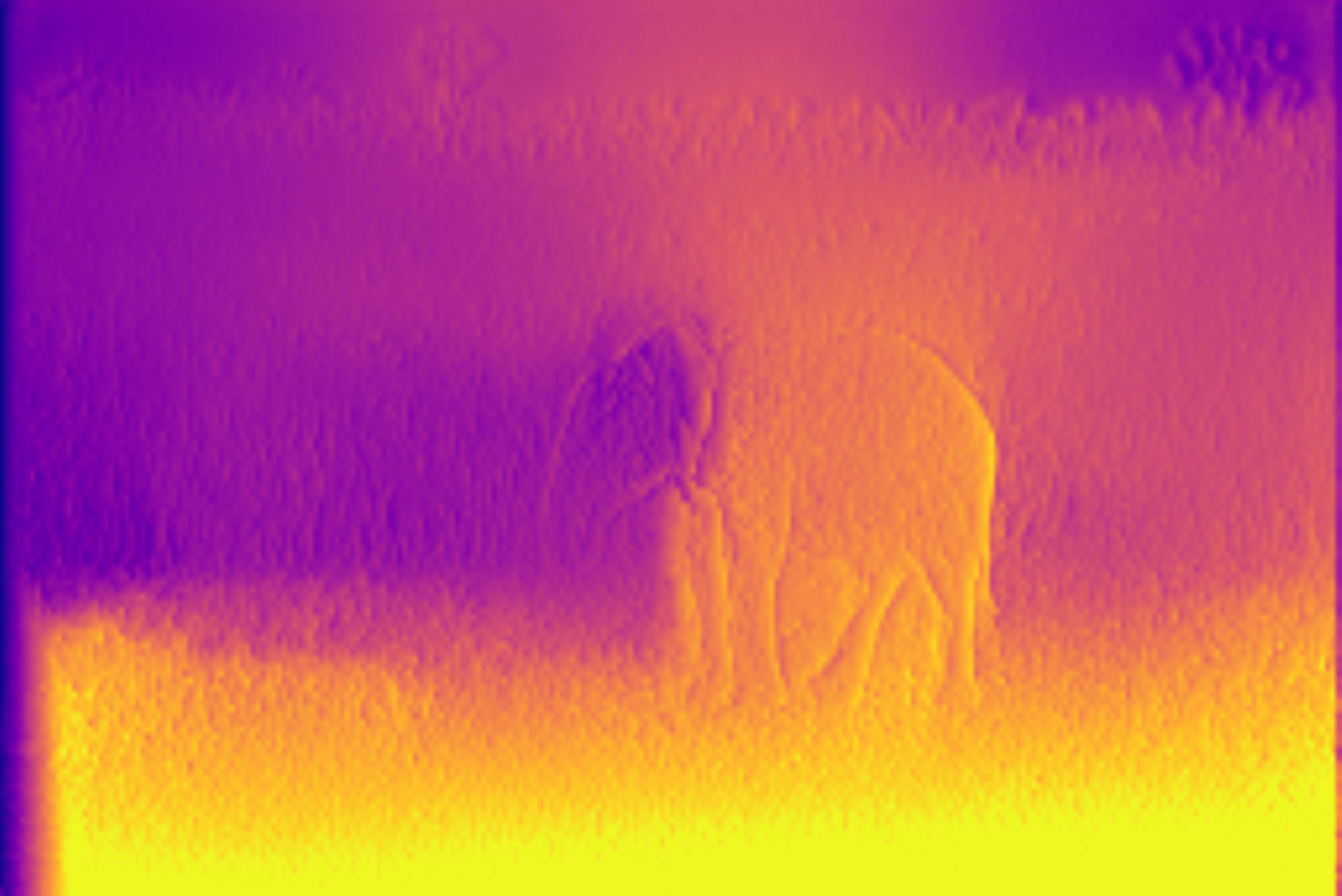}
    \end{subfigure}
    \begin{subfigure}[c]{0.90\columnwidth}
        \centering
        \includegraphics[width=0.30\textwidth]{figs/datasets-examples/5da74266.jpg}
        \includegraphics[width=0.30\textwidth]{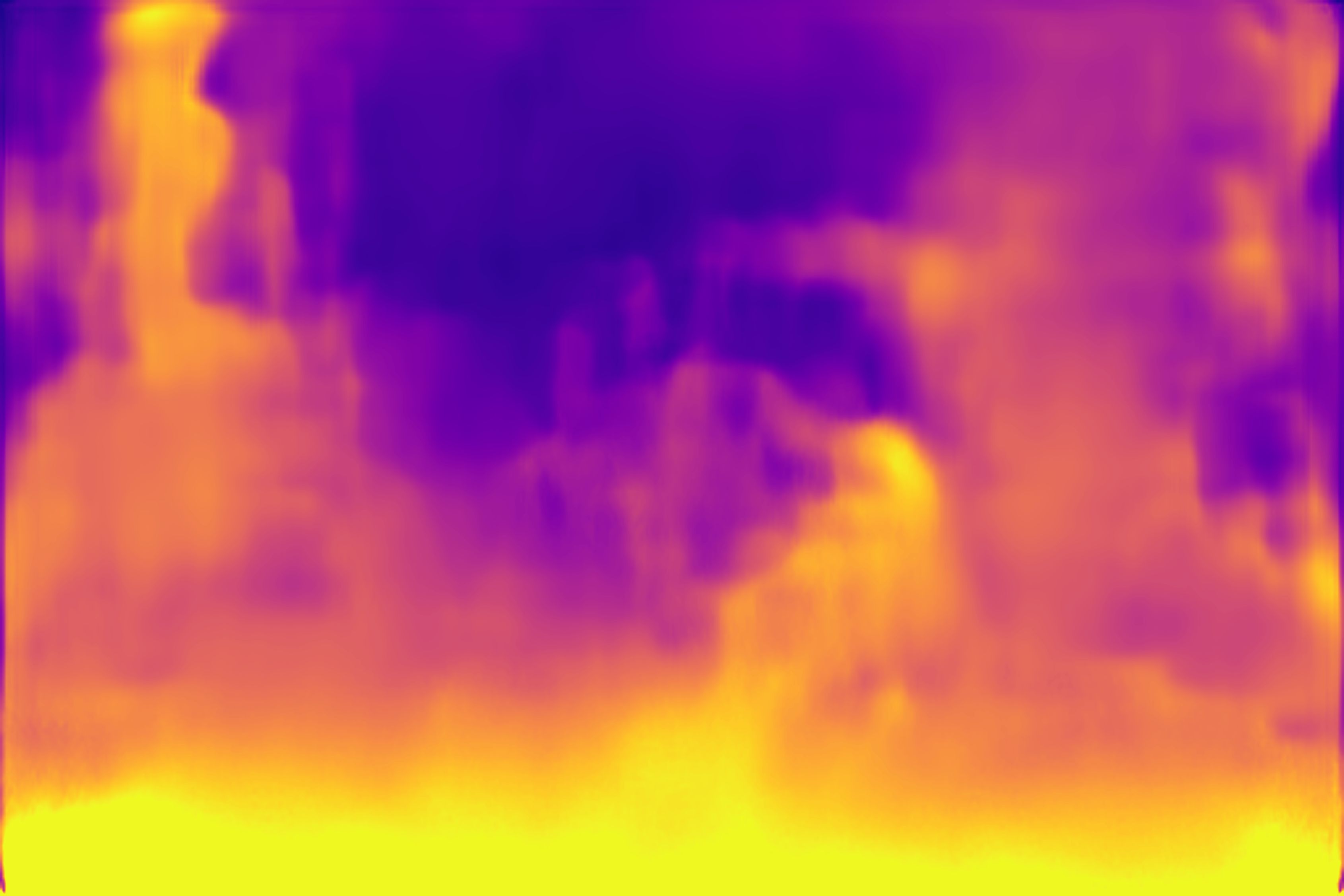}
        \includegraphics[width=0.30\textwidth]{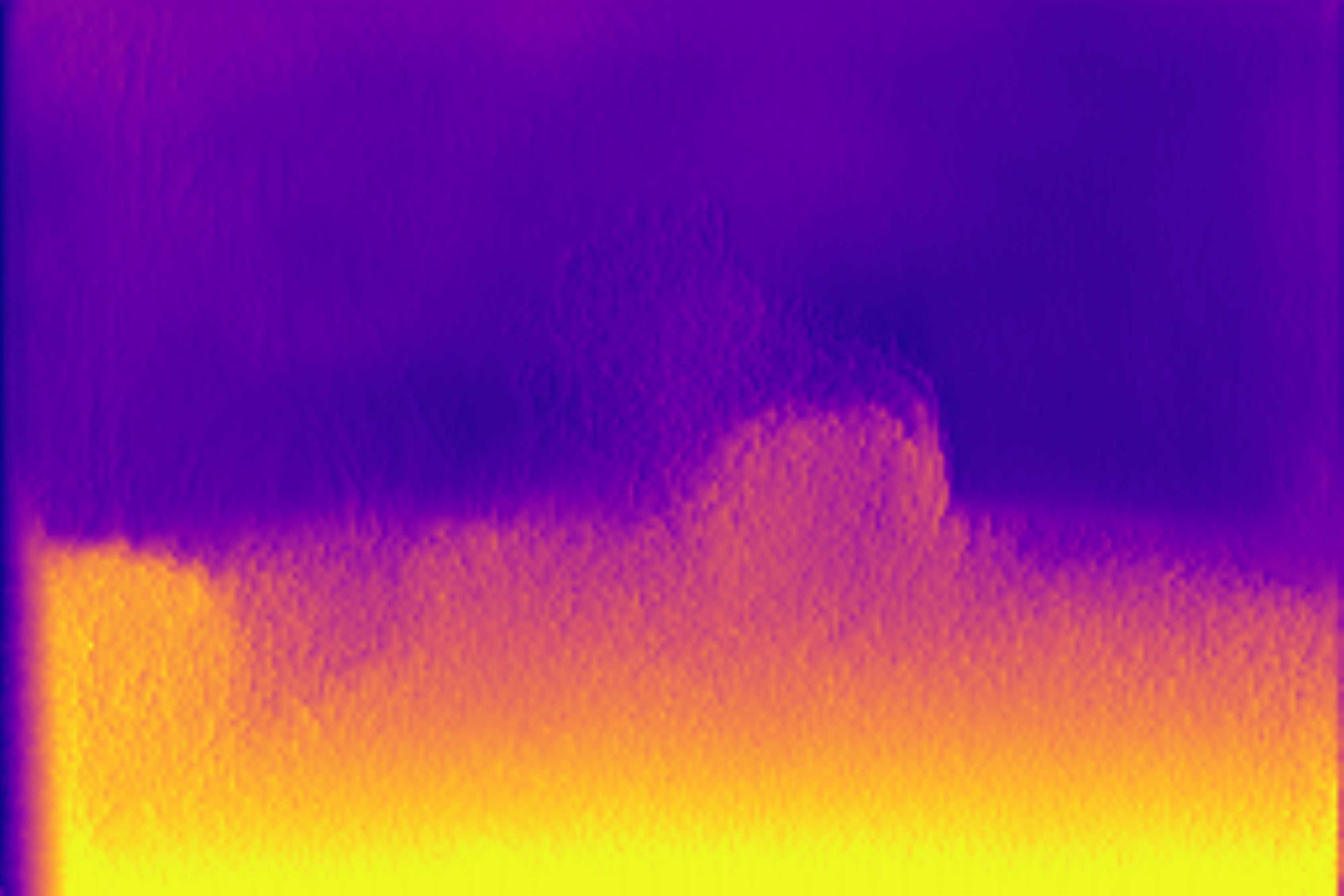}
    \end{subfigure}
\caption{Comparison between methods for depth estimation. The second and third columns illustrate the results obtained by Semantic-Monodepth and PyD-Net networks, respectively.}
\label{fig:compare-depths}
\end{figure}

\begin{figure*}[!t]
    \centering
    \begin{subfigure}[c]{0.7\textwidth}
        \centering
        \includegraphics[width=0.99\textwidth]{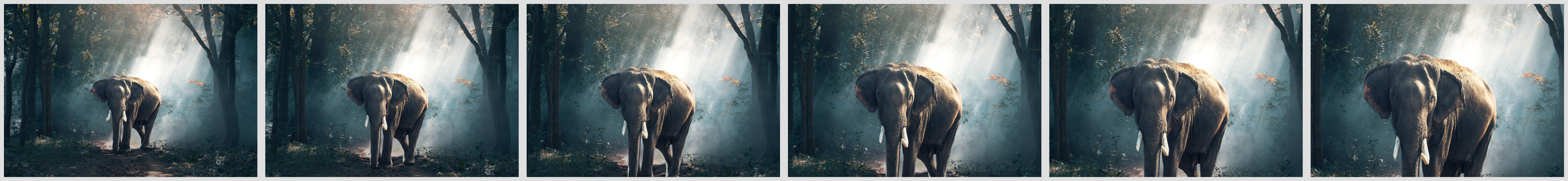}
        \includegraphics[width=0.99\textwidth]{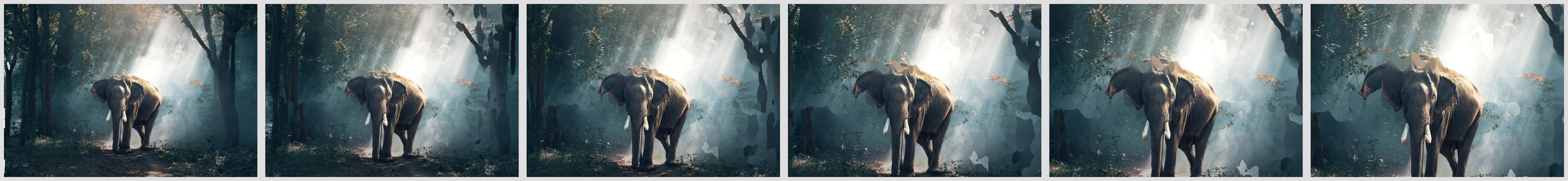}
        \includegraphics[width=0.99\textwidth]{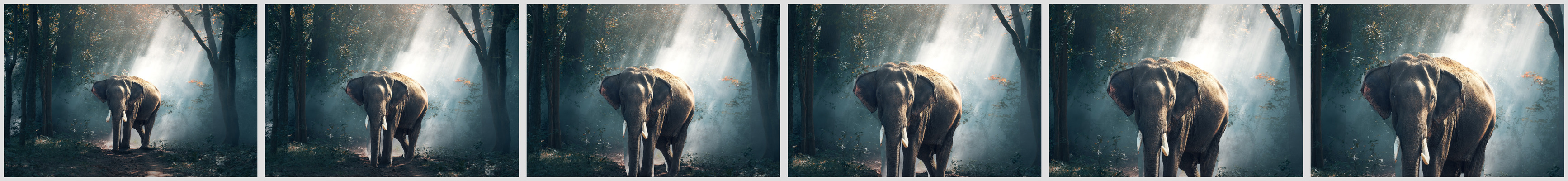}
        \includegraphics[width=0.99\textwidth]{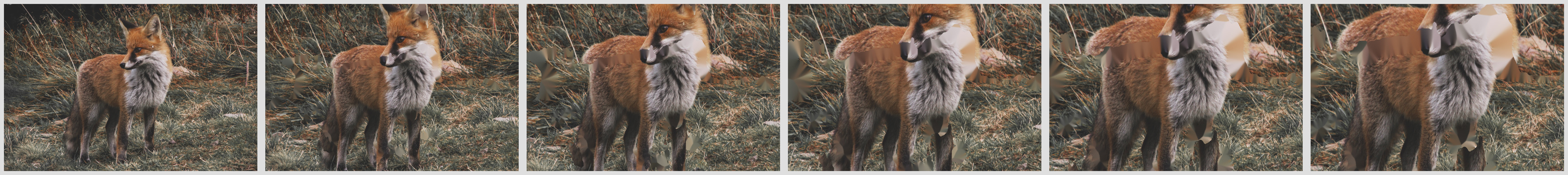}
        \includegraphics[width=0.99\textwidth]{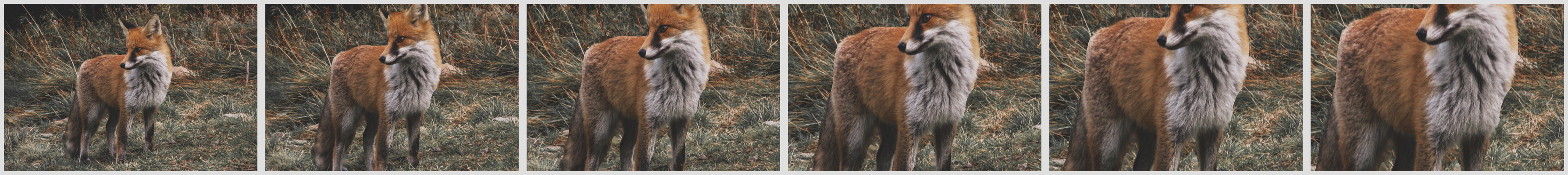}
    \end{subfigure}
    \caption{\review{Examples of segmentation results achieved by segmentation methods. First three rows present the results obtained by the Mask R-CNN (first row), Semantic-Monodepth (second row), and FBNet (third row) networks. The fourh and fifth rows present the results achieved by the Semantic-Monodepth and FBNet networks, respectively. In this example, the Mask R-CNN network was not able to segment the fox.}}
    \label{fig:compare-segm-easy-example-4}
\end{figure*}
\begin{figure*}[!t]
    \centering
    \begin{subfigure}[c]{0.7\textwidth}
        \centering
        \includegraphics[width=0.99\textwidth]{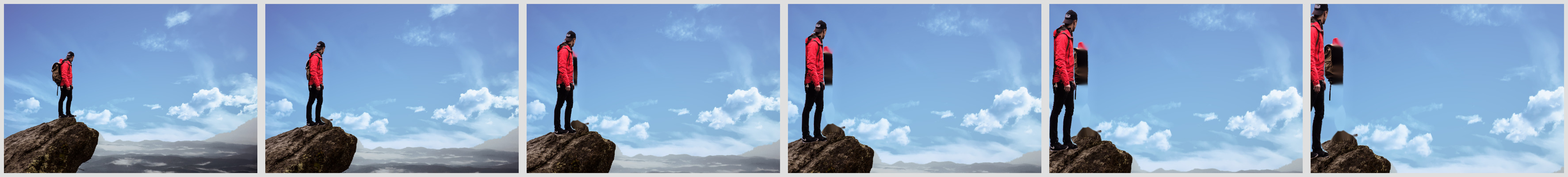}
        \includegraphics[width=0.99\textwidth]{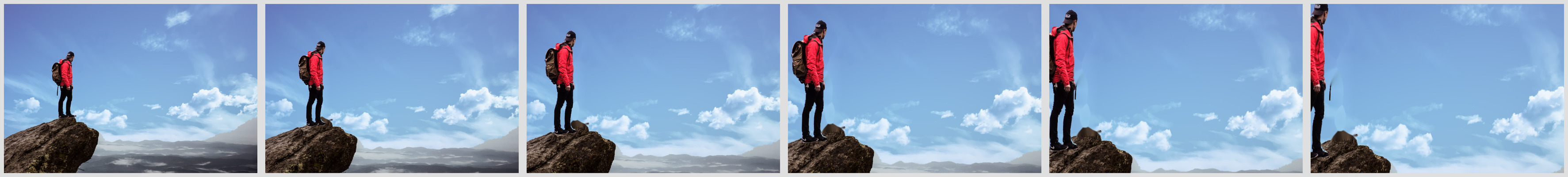}
    \end{subfigure}
    \caption{\review{Example of a parallax motion effect before (top row) and after (bottom row) joining near objects considering their relative distance.}}
    \label{fig:compare-joining-example-1}
\end{figure*}

\vspace*{0.1cm}
\noindent
\textbf{Evaluation Metrics.}
To measure efficiency aspects of our method, we consider both the processing time and the disk usage (in MB). We used the Linux {\em time} command for measuring processing time since this tool can be applied to all evaluated methods, regardless the programming language.
Regarding the efficacy aspects, we performed a visual inspection to measure the quality of a video containing a parallax motion effect due to inherent subjectivity present in this task.\footnote{A supplementary material with more examples and videos containing parallax motion effects generated by our method can be found in \url{https://allansp84.github.io/motion-parallax/} (As of May 2020).}

\subsection{Comparison of Methods for Depth Estimation and Instance Segmentation}
\label{sec:compare_depth_methods}

This section presents the performance results for the PyD-Net and Semantic-Monodepth networks considering the use of models provided by the authors. We measured the effectiveness of these models upon the KITTI dataset, with confirmed the results reported by the authors~\cite{Menze2015CVPR}. Considering efficiency aspects, the Semantic-Monodepth network spent $21.53$ sec./image, whereas the PyD-Net network spent $12.5$ sec./image. Fig.~\ref{fig:compare-depths} presents a comparison among depth maps obtained with Semantic-Monodepth and PyD-Net networks, from which we could observe that both networks were able to detect the object of interest as a foreground object, but also produced depth maps with several inconsistencies.

In the context of parallax motion effect generation, segmentation methods also play a crucial role in the overall quality of parallax videos. We investigated three networks for instance and semantic segmentation that have different requirements in terms of processing system requirements (see Table~\ref{tab:processing_time}).
\begin{table}[!t]
    \setlength{\tabcolsep}{3.0mm}
    \centering
    \caption{Model size (in MB) and latency (in sec./image) of segmentation methods upon the Parallax60 dataset.}
    \label{tab:processing_time}
    \begin{tabular}{llr}
    \toprule
    \textbf{Method} & \textbf{Model Size} & \textbf{Latency}        \\
    \midrule
    Mask R-CNN (ResNet101) & $483.0$ & $26.11$ \\
    Semantic-Monodepth     & $823.8$ & $18.03$ \\
    FBNet                  & $26.70$ & $13.85$ \\
    \bottomrule
    \end{tabular}
\end{table}

\review{Fig.~\ref{fig:compare-segm-easy-example-4} shows visual results achieved with the segmentation methods evaluated in this work. From this experiment, we observed that Mask R-CNN was not able to find any object, for several input images. In total, Mask R-CNN was able to produce at least one mask for $39/60$ images. In turn, the Semantic-Monodepth and FBNet networks produced masks for all images on the Parallax60 dataset. In terms of quality of parallax videos, in general, both Mask R-CNN and FBNet produced better parallax motion effects, in comparison to Semantic-Monodepth network.}

\subsection{\review{Improving Parallax Motion Effects}}
\label{sec:compare_joining_methods}

This section presents two ideas to improve parallax motion effects. The first strategy concerns with joining near objects, according to their relative distance. For all experiments, we considered a maximum relative distance, for merging two objects, up to $20\%$. From the experimental results and visual quality assessments, we observed that poor quality achieved by the segmentation methods is due to the lack of a clear object of interest. In general, these errors occur in images such as natural, landscape, and indoor images. Fig.~\ref{fig:compare-joining-example-1} shows examples 
in which the refinement of foreground and background components (see Sec.~\ref{sec:proposed-method}) improved the visual quality of parallax motion effects significantly.

%% file: 04-conclusions.tex
\section{Conclusions}
\label{sec:conclusions}

This work presented a method for parallax motion effect generation, considering the use of instance segmentation and depth estimation methods. The methods were evaluated in terms of their ability to segment instances towards delimiting objects, and infer distances between objects in the scene, considering landscape and natural images. For the depth estimation task, achieved results suggest that the PyD-Net network provides good depth estimations at an affordable computational cost, in comparison to Semantic-Monodepth network. For the instance segmentation task, the Mask R-CNN presented better qualitative results than all those networks evaluated in this work. However, this network is time consuming and requires about 0.5GB of storage. A low-cost alternative for this task is the FBNet network, which presented similar results at low computational costs, in terms of storage footprints, requiring 30MB of storage. Finally, some future research venues include the combination of efficient depth and instance segmentation networks in a unified architecture in order to have a fast and lightweight model.